\documentclass[conference]{IEEEtran}
\usepackage[utf8]{inputenc}
\usepackage[T1]{fontenc}
\usepackage{amsmath,amssymb}
\usepackage{graphicx}
\usepackage{booktabs}
\usepackage{cite}
\usepackage{hyperref}
\usepackage{url}
\usepackage{microtype}

\IEEEoverridecommandlockouts

\newcommand{\best}[1]{\textbf{#1}}

\newcommand{\runid}[1]{\textbf{\emph{#1}}}

\title{DECOWAM: Decoupled Whole-Body World-Action Model for\\Legged Mobile Manipulation}

\author{
  \IEEEauthorblockN{Siyuan Ma\textsuperscript{*,1}, Boshi Zhang\textsuperscript{*,1},
  Yutian Zhang\textsuperscript{*,2}, Qinglian Wu\textsuperscript{3},\\
  Jiaqi Zhai\textsuperscript{4}, Dong Wei\textsuperscript{\textdagger,4},
  Qiaojun Yu\textsuperscript{\textdagger,2}}
  \IEEEauthorblockA{\textsuperscript{1}Tsinghua University, Beijing, China\\
  \textsuperscript{2}Shanghai Artificial Intelligence Laboratory, Shanghai, China\\
  \textsuperscript{3}Harbin Institute of Technology, Harbin, China\\
  \textsuperscript{4}Hangzhou Yunshenchu Technology Co., Ltd. (DEEP Robotics), Hangzhou, China\\
  \textsuperscript{*}Equal contribution. \textsuperscript{\textdagger}Corresponding authors.}
}

\begin{document}
\maketitle

\begin{abstract}
Mobile manipulation requires a robot to predict how locomotion and arm motion jointly alter future observations and control. Existing world--action models, developed largely for fixed-base platforms, do not explicitly distinguish camera ego-motion from base and arm actions. Here we introduce DECOWAM, a whole-body world--action model that separates these factors through dedicated conditional interfaces. DECOWAM freezes an adapted FastWAM backbone and trains residual adapters, an action-equivalent future bottleneck distilled from privileged observations, adversarially separated base and arm latents, and base-velocity conditioning for video prediction. We further introduce \textsc{ARMDOG}, a real-robot dataset that synchronizes video, whole-body state and action, and language. On a fixed replay protocol, DECOWAM improved both future-video and action prediction over FastWAM, reducing action MSE by 21.7\% with 25.95M trainable adaptation parameters. Across 79 closed-loop trials per method, it achieved the highest observed whole-body coordination and base-displacement robustness among the compared systems, while task completion remained comparable to the strongest baseline. These results show that embodiment-aware factorization can support parameter-efficient joint visual prediction and whole-body control under moving viewpoints.
\end{abstract}

\section{Introduction}

A robot that can both \emph{move through} and \emph{act on} the physical world is the long-standing target of embodied AI research. Recent vision--language--action (VLA) and world--action models---Aloha~\cite{zhao2023aloha}, RT-1/2~\cite{brohan2022rt1,brohan2023rt2}, OpenVLA~\cite{kim2024openvla}, PaLM-E~\cite{driess2023palme}, SayCan~\cite{ahn2022saycan}, $\pi_0$~\cite{black2024pi0}, RDT-2~\cite{liu2024rdt}, and Motus---have produced impressive results on stationary bimanual platforms, where the camera geometry is fixed and the action space is restricted to arm trajectories. Legged mobile manipulators change the modeling problem: the camera is carried by a moving base, and the policy must coordinate high-rate arm motion with lower-rate base velocity commands.

\begin{figure*}[!t]
  \centering
  \includegraphics[width=0.98\textwidth]{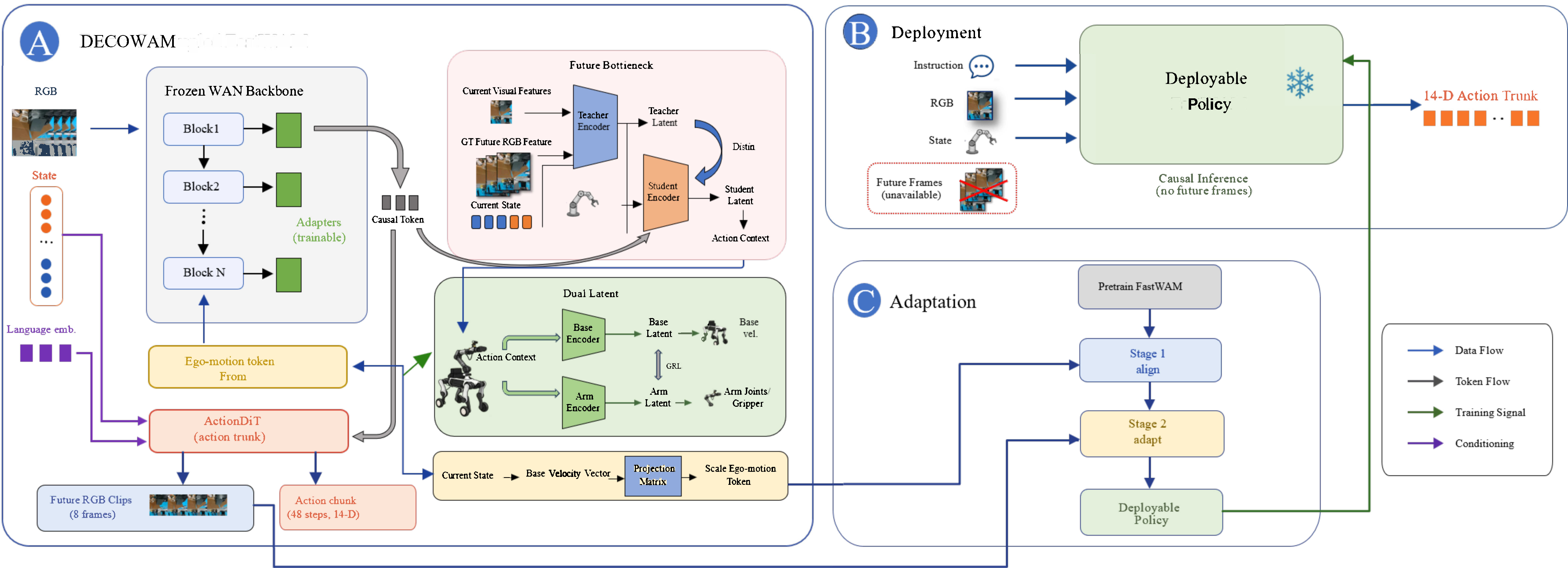}
  \caption{DECOWAM architecture, training, and deployment. (A) A frozen WAN backbone is augmented with trainable residual adapters, a teacher--student future bottleneck, separated base/arm latents, and base-velocity ego-motion conditioning; ActionDiT produces future RGB clips and a 48-step, 14-D action chunk. (B) Deployment removes future-frame inputs and the privileged teacher path, yielding strictly causal 14-D action inference. (C) Staged training first aligns pretrained FastWAM to \textsc{ARMDOG} and then adapts the decoupled modules to obtain the deployable policy.}
  \label{fig:intro_overview}
\end{figure*}

\paragraph{Problem formulation.}
Given an instruction $\ell$, a current RGB observation $x_0$, and robot state $s$, a whole-body world--action model predicts a future action chunk $\hat{\mathbf{a}}_{1:K}$ and a future video clip $\hat{x}_{1:T}$. In our setting, $\mathbf{a}_{1:K}\in\mathbb{R}^{K\times14}$ contains arm joints, gripper state, base velocity, and loader-compatible padding. The difficulty is structural. First, the camera coordinate system changes with the legged base, so apparent image motion mixes scene dynamics, arm motion, and ego-motion. Second, the action vector is multi-factor: arm/gripper channels and base-velocity channels have different semantics and different control time scales. Third, base velocity has a dual role: it is a target to be predicted by the action expert and an observation that explains camera motion for the video expert.

\paragraph{Why is legged mobile manipulation harder?}
These properties make legged-arm modeling different from fixed-base manipulation:
\begin{itemize}
  \item \textbf{Dynamic viewpoint.} The on-board camera moves with the base. Hand--eye geometry varies continuously, and image streams contain a mixture of ego-motion and scene motion. A mobile-manipulation world model needs a route for representing camera motion rather than treating all pixel displacement as scene dynamics.
  \item \textbf{Multi-rate action coupling.} Arm joint trajectories require high-rate control ($\sim$15--30\,Hz), while base velocity commands are typically issued at a lower rate ($\sim$3--5\,Hz). Concatenating both into a single uniformly sampled action chunk asks one representation to cover navigation-scale velocity and manipulation-scale joint corrections at the same time.
  \item \textbf{Hierarchical intent.} Real tasks interleave \emph{where to go} with \emph{how to act}. A single monolithic latent struggles to represent navigation-scale decisions and manipulation-scale corrections simultaneously.
\end{itemize}

\paragraph{Our approach.}
The central idea of this paper is a decoupled modeling paradigm for mobile manipulation: a world--action model should represent \emph{where the base moves}, \emph{how the arm acts}, and \emph{how the camera ego-motion changes future pixels} as explicit factors. We implement this paradigm on top of FastWAM, a Wan-2.2-based world--action backbone with a paired ActionDiT branch for action chunks. The action-equivalent future bottleneck supplies a modest causal training signal from privileged future latents. Staged frozen adaptation turns the method into a parameter-efficient system by keeping the base FastWAM prior fixed in the final stage and learning only residual robot-specific pathways. The base/arm dual latent with GRL is the main action-factorization mechanism, separating navigation-scale base commands from manipulation-scale arm commands. The base-velocity token is the explicit ego-motion interface for the video branch: it conditions future visual rollout on the current normalized base velocity rather than asking the video branch to infer camera motion only from pixels.

\paragraph{Dataset.}
Method development is only half of the story. Legged mobile manipulation requires data in which visual change, base ego-motion, arm motion, and language intent are synchronized rather than recorded as separate logs. We therefore build \textsc{ARMDOG}, a real-robot data resource for a quadrupedal platform with a 6-DoF arm. Its contribution is the embodiment-complete model interface: each converted episode aligns a 15\,Hz RGB video stream, a $T{\times}14$ whole-body state/action tensor with explicit base and arm channels, natural-language instruction text, and a precomputed language embedding. The current converted world--action snapshot contains 217 episodes from 27 task folders and 56,041 synchronized frames. This data organization is what makes it possible to train and evaluate base/arm factorization, ego-motion-aware video conditioning, and whole-body world--action prediction within one replay protocol.

\paragraph{Contributions.}
\begin{enumerate}
  \item We formulate legged mobile manipulation as a decoupled world--action modeling problem in which base control, arm manipulation, and camera ego-motion enter through explicit, semantically aligned interfaces.
  \item We realize this formulation in DECOWAM through frozen residual adaptation, causal distillation from privileged future latents, adversarial base/arm factorization, and base-velocity-conditioned video prediction.
  \item We introduce the synchronized \textsc{ARMDOG} dataset and evaluate DECOWAM through controlled replay and real-robot experiments, demonstrating parameter-efficient prediction, improved whole-body coordination, and stronger perturbation tolerance.
\end{enumerate}

\section{Related Work}

\paragraph{Vision--language--action models for fixed-base robots.}

Modern VLA models---RT-1/2~\cite{brohan2022rt1,brohan2023rt2}, OpenVLA~\cite{kim2024openvla}, $\pi_0$~\cite{black2024pi0}, Octo~\cite{octo2024}, RDT-2~\cite{liu2024rdt}, X-VLA~\cite{xvla2025}, PaLM-E~\cite{driess2023palme}, SayCan~\cite{ahn2022saycan}, Gato~\cite{reed2022gato}, and RoboCat~\cite{bousmalis2023robocat}---map language and pixels to robot actions through transformer policies pre-trained on large heterogeneous corpora. Task-conditioned manipulation policies such as BC-Z, PerAct, VIMA, and Diffusion Policy~\cite{jang2021bcz,shridhar2022peract,jiang2023vima,chi2023diffusion} provide complementary evidence that language, visual tokens, and action trajectories can share a policy interface. These models provide strong action-policy references, and we include adapted $\pi_{0.5}$ and X-VLA runs in the replay experiments. Their standard action heads do not explicitly model future RGB rollout or separate base-induced camera motion from arm-induced scene change.

\paragraph{World models for robotic control.}

World-model approaches~\cite{ha2018world,hafner2021dreamerv2,hafner2023dreamer,hansen2024tdmpc2,bruce2024genie} generate future observations and use them either as a learned simulator or as a structural prior on the policy. Video diffusion and interactive-video systems~\cite{ho2022videodiffusion,brooks2024sora,yang2024unisim} further show that future visual prediction can serve as a powerful generative prior. Recent video-based world--action models such as UniSim~\cite{yang2024unisim}, UVA~\cite{uva2025}, X-WAM~\cite{xwam2026}, FastWAM, and Motus combine visual prediction with action modeling. Our work does not pursue general simulator scaling. It studies embodiment-specific adaptation: how to keep a pretrained video prior useful while separating ego-motion, base action, arm action, and future-equivalent control information on a legged-manipulation dataset.

\paragraph{Locomotion and mobile manipulation.}

Legged locomotion has been studied extensively through reinforcement learning~\cite{rudin2022learning,margolis2023walk}, with controllers that output base velocity commands but do not engage manipulation. Wheeled and home mobile-manipulation systems~\cite{yokoyama2023adaptive,xia2021relmogen,spot2023,homerobot2023,szot2021habitat2,srivastava2022behavior} have produced cascaded policies that separate navigation and manipulation. Mobile ALOHA~\cite{fu2024mobilealoha} demonstrates the value of whole-body teleoperation for bimanual mobile manipulation, but few works learn a unified policy, and fewer still ground that policy in a predictive video model. We are not aware of prior work that performs unified whole-body world-action modeling on a legged base with a manipulator.

\paragraph{Robot datasets.}

Open-X Embodiment~\cite{openx2023}, DROID~\cite{khazatsky2024droid}, BridgeData V2~\cite{walke2023bridgedatav2}, RoboNet~\cite{dasari2019robonet}, RLBench~\cite{james2020rlbench}, CALVIN~\cite{mees2022calvin}, Language Table~\cite{lynch2023languagetable}, ManiSkill2~\cite{gu2023maniskill2}, LIBERO~\cite{liu2023libero}, RoboTwin~\cite{robotwin2025}, and Aloha-derived collections~\cite{zhao2023aloha} provide broad coverage for fixed-base or wheeled manipulation, but they do not expose the combination needed by our model: a legged-base ego-motion stream, manipulator actions, visual observations, and language instructions in one synchronized training unit. \textsc{ARMDOG} fills this complementary regime by recording quadrupedal mobile manipulation and converting it into a video--state/action--language format consumed by FastWAM-style world--action models.

\section{Background: FastWAM World--Action Backbone}

We briefly review the FastWAM architecture, on which our method builds. FastWAM couples a Wan-2.2 video diffusion backbone~\cite{wan2024} with an ActionDiT branch that predicts chunked actions using the same continuous-flow training interface. In the \textsc{ARMDOG} configuration, the model consumes the current RGB frame, a precomputed language context, and a 14-D proprioceptive state token, then jointly produces future RGB frames and a 48-step whole-body action chunk. The original FastWAM checkpoint evaluated here has 6725.44M total parameters and 6020.75M trainable parameters during full fine-tuning.

We retain this joint video/action factorization and add four mechanisms that decouple the adaptation problem in parameter space, future-information usage, action factors, and ego-motion conditioning.

\section{Method}

\subsection{Problem Formulation}
\label{sec:overview}

Let $\ell$ denote a language instruction, $x_0$ the current RGB observation, and $s_0\in\mathbb{R}^{14}$ the current whole-body state. We learn the conditional joint model
\begin{equation}
  p_\theta\!\left(x_{1:T},\mathbf{a}_{1:K}\mid x_0,s_0,\ell\right),
  \label{eq:joint_problem}
\end{equation}
where $x_{1:T}$ is a future video and $\mathbf{a}_{1:K}\in\mathbb{R}^{K\times14}$ is an action chunk. We use $T{=}8$ and $K{=}48$. Each action has the semantic decomposition
\begin{equation}
  \mathbf{a}_{k}
  =
  \left[
    \mathbf{a}^{\mathrm{arm}}_{k},
    a^{\mathrm{grip}}_{k},
    \mathbf{a}^{\mathrm{base}}_{k},
    \mathbf{a}^{\mathrm{pad}}_{k}
  \right],
  \qquad
  \mathbf{a}^{\mathrm{base}}_{k}\in\mathbb{R}^{3}.
  \label{eq:action_partition}
\end{equation}
The arm and gripper occupy channels $[0{:}7]$, base velocity occupies $[7{:}10]$, and loader padding occupies $[10{:}14]$. Base velocity is both a control target and a source of camera ego-motion. DECOWAM therefore separates arm control, base control, and visual ego-motion instead of encoding them in one undifferentiated context.

Our backbone is FastWAM, which pairs an ActionDiT action expert with a WAN video expert. Both experts receive language and proprioceptive context. The action expert predicts flow over action chunks, while the video expert predicts flow over future visual latents. DECOWAM preserves this joint interface and changes how embodiment-specific information conditions each expert. Future observations supervise the model only during training, so deployment remains causal in $(x_0,s_0,\ell)$.

\subsection{Staged Parameter-Efficient Adaptation}
\label{sec:staged}

Training separates domain alignment from structural adaptation. In Stage 1, all FastWAM parameters $\Theta$ are adapted to \textsc{ARMDOG} for 50k steps:
\begin{equation}
  \Theta^{(1)}
  =
  \arg\min_{\Theta}
  \mathbb{E}_{\mathcal D}
  \left[
    \mathcal L_{\mathrm{video}}(\Theta)
    +
    \mathcal L_{\mathrm{action}}(\Theta)
  \right].
  \label{eq:stage1}
\end{equation}
This stage aligns the video prior, action expert, and proprioceptive interface with the moving-camera observations and quadruped--arm action space.

Stage 2 freezes $\Theta^{(1)}$ and optimizes only
\begin{equation}
  \Phi
  =
  \left\{
    \phi_{\mathrm{adp}},
    \phi_q,
    \phi_{\mathrm{ba}},
    \phi_{\mathrm{ego}}
  \right\},
  \qquad
  \Phi^\star
  =
  \arg\min_{\Phi}
  \mathcal L\!\left(\Theta^{(1)},\Phi\right).
  \label{eq:stage2_params}
\end{equation}
The four parameter groups represent residual adapters, an action-equivalent future bottleneck, base/arm factorization, and ego-motion conditioning. This restriction reduces the Stage-2 trainable footprint from 6020.75M to 25.95M parameters.

The frozen WAN backbone is adapted after each block through
\begin{equation}
  h_l^{+}
  =
  h_l
  +
  \alpha_l W_{\mathrm{up}}^{(l)}
  \sigma\!\left(
    W_{\mathrm{down}}^{(l)}\mathrm{LN}(h_l)
  \right),
  \label{eq:adapter}
\end{equation}
where $W_{\mathrm{down}}^{(l)}$ projects to a 128-D bottleneck, $W_{\mathrm{up}}^{(l)}$ restores the hidden dimension, and $\sigma$ is SiLU. The residual branch learns a compact, robot-specific correction while preserving the pretrained video prior.

\subsection{Decoupled Conditional Interfaces}
\label{sec:adapter}

\paragraph{Action-equivalent future bottleneck.}
\label{sec:quotient}

Future frames contain information about action-equivalent outcomes that is unavailable from the current frame alone. We distill this privileged information from a teacher into a causal student.

\begin{figure}[t]
  \centering
  \includegraphics[width=0.98\columnwidth]{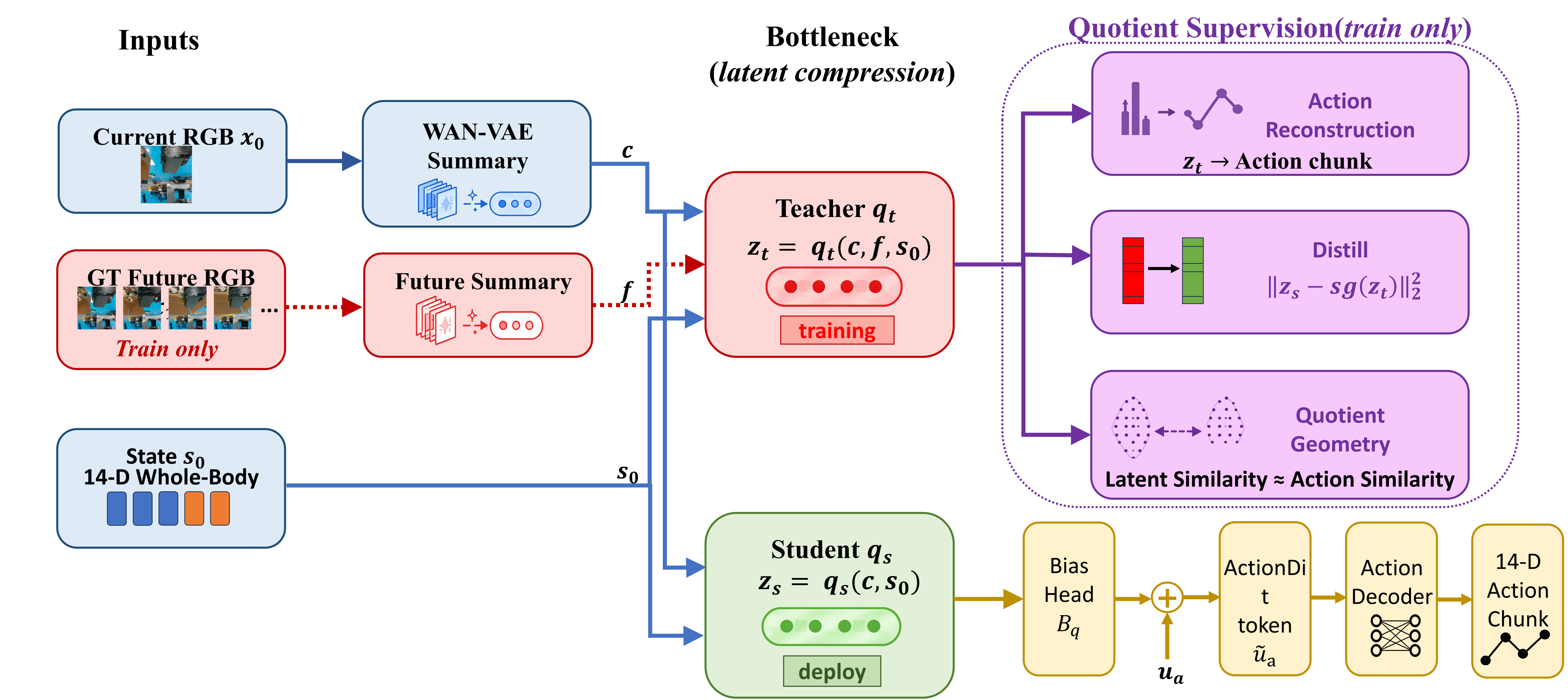}
  \caption{Future-information bottleneck. A privileged teacher observes current and future visual summaries, whereas the causal student observes only the current summary and robot state. Only the student is retained during deployment.}
  \label{fig:quotient_method}
\end{figure}

Let $e_0=\psi_{\mathrm{vae}}(x_0)$ and $e_{1:T}=\psi_{\mathrm{vae}}(x_{1:T})$ be WAN-VAE latents. We summarize each latent tensor using
\begin{equation}
  c=\rho(e_0),
  \qquad
  f=\rho(e_{1:T}),
  \qquad
  \rho(e)=\left[\operatorname{mean}(e),\operatorname{std}(e)\right].
  \label{eq:vae_summary}
\end{equation}
The teacher and student embeddings are
\begin{equation}
  z_t=q_t([c,f,s_0]),
  \qquad
  z_s=q_s([c,s_0]),
  \qquad
  z_t,z_s\in\mathbb{R}^{d_q}.
  \label{eq:quotient_latents}
\end{equation}
Only $z_s$ conditions the causal action expert:
\begin{equation}
  \tilde{u}^{a}=u^{a}+\eta_q B_q z_s,
  \label{eq:quotient_bias}
\end{equation}
where $u^a$ denotes its context tokens and $\eta_q\in[0,1]$ controls the residual bias.

The bottleneck is trained with action reconstruction, teacher--student distillation, and geometry preservation:
\begin{align}
  \mathcal L_{\mathrm{rec}}^q
  ={}&
  \left\|r_s(z_s)-\mathbf a_{1:K}\right\|_2^2
  \nonumber\\
  &+
  \left\|r_t(z_t)-\mathbf a_{1:K}\right\|_2^2,
  \nonumber\\
  \mathcal L_q
  ={}&
  \lambda_{\mathrm{act}}^q\mathcal L_{\mathrm{rec}}^q
  +
  \lambda_{\mathrm{dist}}^q
  \left\|z_s-\operatorname{sg}(z_t)\right\|_2^2
  \nonumber\\
  &+
  \lambda_{\mathrm{geom}}^q
  \mathcal L_{\mathrm{geom}},
  \label{eq:quotient}
\end{align}
where $\operatorname{sg}$ stops gradients. For a batch of size $B$, the geometry term is
\begin{equation}
  d_z^{ij}=\frac{\|z_t^i-z_t^j\|_2}{\tau_z},
  \qquad
  d_a^{ij}=\frac{\|\mathbf a^i_{1:K}-\mathbf a^j_{1:K}\|_2}{\tau_a},
  \label{eq:quotient_distances}
\end{equation}
\begin{equation}
  \begin{aligned}
    \mathcal L_{\mathrm{geom}}
    &={}
    \bigl(B(B-1)\bigr)^{-1}
    \\
    &\quad\times
    \sum_{i\neq j}
    \operatorname{SL1}\!\left(d_z^{ij},d_a^{ij}\right).
  \end{aligned}
  \label{eq:quotient_geom}
\end{equation}
with robust scales $\tau_z$ and $\tau_a$ obtained from batch medians. Thus, trajectories with similar normalized actions are encouraged to remain close in the privileged latent space. Distillation transfers this structure to the deployable student.

\paragraph{Base--arm factorization.}
\label{sec:dual}

The action context contains navigation-scale and manipulation-scale information. We map its pooled representation into two 16-D factors:
\begin{equation}
  z_{\mathrm{base}}=b_{\phi}(u^a),
  \qquad
  z_{\mathrm{arm}}=m_{\phi}(u^a),
  \qquad
  z_{\mathrm{base}},z_{\mathrm{arm}}\in\mathbb{R}^{16}.
  \label{eq:dual_latents}
\end{equation}
Their concatenation conditions the action expert through
\begin{equation}
  \bar{u}^{a}
  =
  \tilde{u}^{a}
  +
  \eta_{\mathrm{ba}}B_{\mathrm{ba}}
  [z_{\mathrm{base}},z_{\mathrm{arm}}].
  \label{eq:dual_bias}
\end{equation}
Let $\mathbf a^b_{1:K}=\mathbf a_{1:K,7:10}$ and $\mathbf a^m_{1:K}=\mathbf a_{1:K,0:7}$. Direct heads preserve the assigned factor, while gradient-reversal cross heads suppress information about the opposite factor:
\begin{align}
  \mathcal L_{\mathrm{disent}}
  ={}&
  \left\|g_b(z_{\mathrm{base}})-\mathbf a^b_{1:K}\right\|_2^2
  +
  \left\|g_m(z_{\mathrm{arm}})-\mathbf a^m_{1:K}\right\|_2^2
  \nonumber\\
  &+
  \left\|\tilde g_b(\operatorname{GRL}(z_{\mathrm{arm}}))-\mathbf a^b_{1:K}\right\|_2^2
  \nonumber\\
  &+
  \left\|\tilde g_m(\operatorname{GRL}(z_{\mathrm{base}}))-\mathbf a^m_{1:K}\right\|_2^2.
  \label{eq:disent}
\end{align}
The prediction heads minimize all reconstruction terms, whereas gradient reversal changes the sign of cross-task gradients entering the encoders. Each latent is therefore encouraged to retain its assigned control factor and discard the other.

\paragraph{Ego-motion-aware video conditioning.}
\label{sec:basevel}

For a body-mounted camera, apparent image motion combines scene dynamics, manipulator motion, and base-induced viewpoint change. We expose the last component using the normalized current base velocity
\begin{equation}
  v_0=\Pi_{\mathrm{base}}(s_0)=(v_x,v_y,\omega_z)\in\mathbb{R}^{3}.
  \label{eq:base_velocity}
\end{equation}
Each video token receives the same projected ego-motion condition:
\begin{equation}
  \tilde h_i^v=h_i^v+\beta B_vv_0,
  \qquad i=1,\ldots,N_v,
  \label{eq:basevel}
\end{equation}
where $B_v$ maps velocity into the WAN hidden dimension. This token does not impose geometric warping. It supplies an explicit explanatory variable for camera-frame motion. Base velocity consequently acts as an action target in Eq.~\eqref{eq:action_partition} and a visual condition in Eq.~\eqref{eq:basevel}.

\subsection{Training Objective and Deployment}
\label{sec:loss}

Both experts use conditional flow matching~\cite{lipman2023flow}. For a target $y\in\{\mathbf a_{1:K},e_{1:T}\}$, noise $\epsilon\sim\mathcal N(0,I)$, and time $\tau\sim\mathcal U(0,1)$, define
\begin{equation}
  y_\tau=(1-\tau)\epsilon+\tau y,
  \qquad
  v^\star(y_\tau,\tau)=y-\epsilon.
  \label{eq:flow_target}
\end{equation}
The corresponding objective is
\begin{equation}
  \mathcal L_{\mathrm{FM}}(F_\theta;y,c)
  =
  \mathbb E_{\tau,\epsilon}
  \left[
    \left\|F_\theta(y_\tau,\tau,c)-v^\star(y_\tau,\tau)\right\|_2^2
  \right].
  \label{eq:fm_loss}
\end{equation}
We instantiate this loss as $\mathcal L_{\mathrm{action}}$ with context $\bar u^a$ and as $\mathcal L_{\mathrm{video}}$ with context $\tilde h^v$. The complete Stage-2 objective is
\begin{equation}
  \mathcal L
  =
  \lambda_v\mathcal L_{\mathrm{video}}
  +
  \lambda_a\mathcal L_{\mathrm{action}}
  +
  \gamma_q\lambda_q\mathcal L_q
  +
  \gamma_{\mathrm{ba}}\lambda_{\mathrm{ba}}\mathcal L_{\mathrm{disent}}.
  \label{eq:loss}
\end{equation}
We set $\lambda_v=\lambda_a=1.0$, $\lambda_q=0.2$, and $\lambda_{\mathrm{ba}}=0.1$. The reported run sets $\gamma_q$, $\gamma_{\mathrm{ba}}$, $\eta_q$, and $\eta_{\mathrm{ba}}$ to one throughout Stage 2.

At inference, the teacher $q_t$ and all auxiliary prediction heads are removed. The model computes $z_s$, $(z_{\mathrm{base}},z_{\mathrm{arm}})$, and $v_0$ from current inputs, then samples both flows using only $(x_0,s_0,\ell)$.

\section{The \textsc{ARMDOG} Dataset}
\label{sec:dataset}

\textsc{ARMDOG} is a model-facing real-robot resource for legged mobile manipulation. It synchronizes moving-camera RGB-D, proprioception, IMU, base state, whole-body commands, and language instructions on a wheeled quadruped robot with 16 leg joints, a 6-DoF arm, and a 1-DoF gripper. The FastWAM-compatible conversion aligns raw HDF5 streams to 15\,Hz and stores each episode as $e_i=(V_i,Q_i,\ell_i,\phi(\ell_i))$: RGB video, a structured $T_i{\times}14$ state/action tensor, instruction text, and a precomputed language embedding. Training consumes the current frame, eight future frames at $384{\times}320$, the initial state, and a normalized 48-step action chunk.

\begin{figure}[t]
  \centering
  \includegraphics[width=\columnwidth]{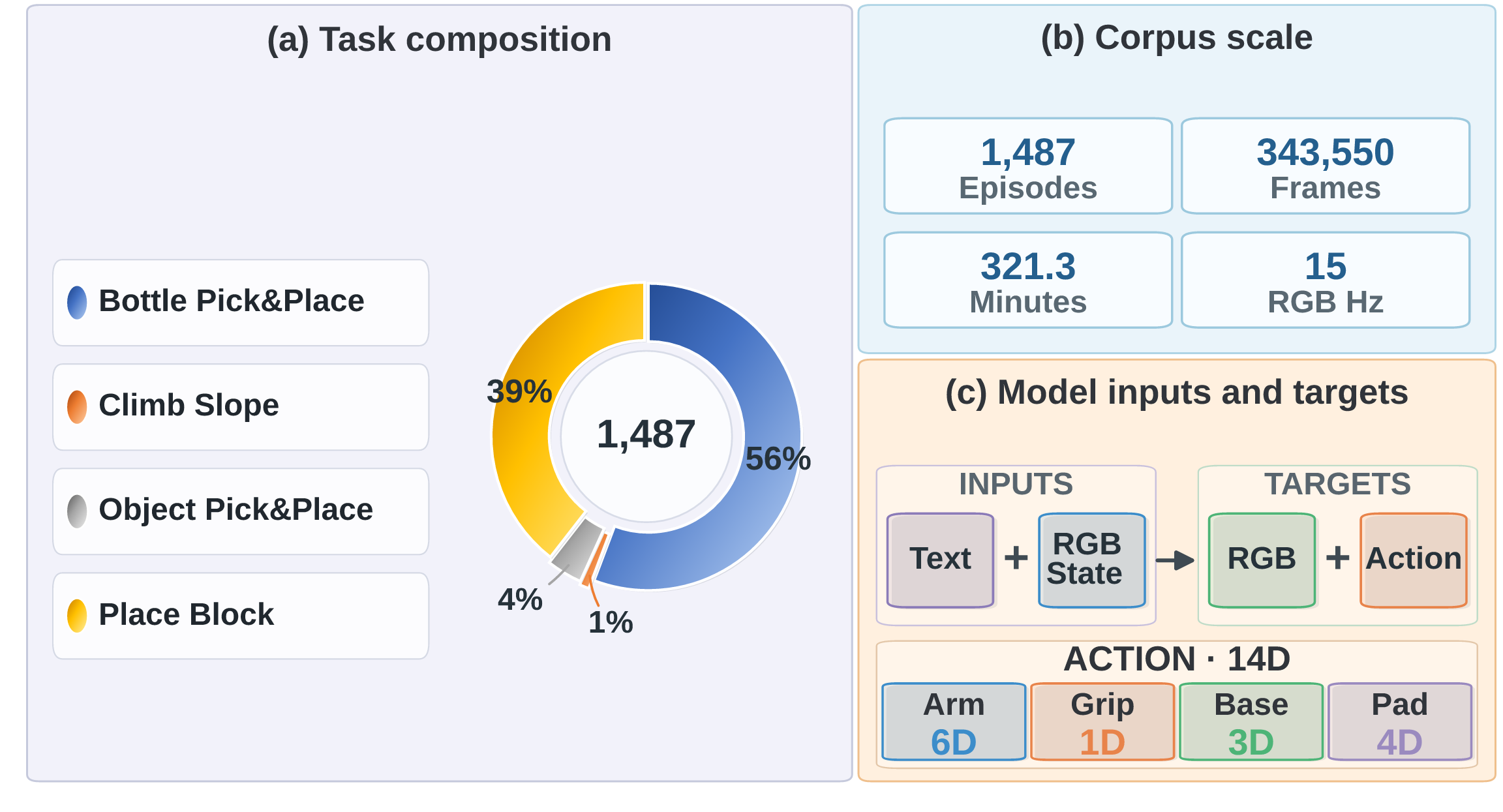}
  \caption{Composition and model-facing structure of \textsc{ARMDOG}. (a) Episode distribution across four task families in the quality-filtered corpus: Bottle Pick\&Place (56\%), Place Block (39\%), Object Pick\&Place (4\%), and Climb Slope (1\%). (b) Corpus scale after timestamp alignment and quality filtering: 1,487 episodes, 343,550 RGB frames, 321.3 minutes, and a 15\,Hz frame rate. (c) Model interface: text, current RGB, and robot state form the inputs, while future RGB and a 14-D action are prediction targets. The action vector separates 6-D arm, 1-D gripper, 3-D base-velocity, and 4-D loader-padding channels.}
  \label{fig:armdog_dataset}
\end{figure}

Figure~\ref{fig:armdog_dataset} summarizes the dataset at the task, corpus, and model-interface levels. In the full quality-filtered corpus, Bottle Pick\&Place and Place Block account for 56\% and 39\% of episodes, respectively, while Object Pick\&Place contributes 4\% and Climb Slope contributes 1\%. This mix emphasizes object-centric mobile manipulation while retaining a smaller locomotion-focused slice. Panel (c) makes the learning contract explicit: language text together with current RGB and state conditions future RGB and structured whole-body action prediction. The corpus-level counts in the figure are distinct from the model-specific training and evaluation subsets described below.

The 14-D tensor exposes the embodiment factors used by our method: indices $[0{:}6]$ are arm targets, $[6]$ is gripper opening, $[7{:}10]$ is base velocity $(v_x,v_y,\omega_z)$, and $[10{:}14]$ is loader padding. The June 5, 2026 audit starts from 795 raw HDF5 episodes, quarantines duplicates, aligns streams by timestamp interpolation, repairs short dropouts, and records per-episode integrity metadata. As reported in Fig.~\ref{fig:armdog_dataset}(b), the dataset after timestamp alignment and quality filtering contains 1487 episodes from 5 task folders and 343,550 frames, approximately 321.3 minutes at 15\,Hz. These full-corpus statistics are distinct from the downstream subsets: frozen decoupled Stage-2 uses 214 episodes from 26 tasks after excluding Legacy val, while all replay results use the fixed Box-val slice with 23 episodes, eight tasks, and 4,323 frames. The planned release includes raw/cleaned HDF5 files, the world--action conversion, immutable split manifests, and a datasheet~\cite{gebru2021datasheets}.

\section{Experiments}
\label{sec:exp}

The experiments follow an evidence ladder from controlled prediction to physical deployment. We first compare DECOWAM with its FastWAM initialization, then test the structured paths through ablation. Broader reference comparisons establish its position among VLA and WAM systems. Closed-loop trials finally assess task progress, whole-body coordination, and robustness on the physical robot.

\subsection{Replay Protocol and Compared Models}

Open-loop replay uses a fixed 23-episode \runid{box\_val\_\_*} slice drawn from eight box-manipulation task folders. The slice covers \runid{box\_move}, \runid{box\_soft}, and \runid{box\_stay} variants. FastWAM-family models receive identical current-frame, state, and language inputs. Each model predicts eight future RGB frames at $384\times320$ resolution and a 48-step, 14-D action chunk. Unless stated otherwise, results are computed over the same 16 replay batches with shared normalization, inputs, and evaluator code.

\paragraph{Models.}
The primary baseline is the Stage-1 FastWAM model trained on the converted \textsc{ARMDOG} snapshot. Checkpoints at 40k, 50k, and 80k steps expose sensitivity to the stopping point. DECOWAM starts from the 50k checkpoint and freezes all FastWAM parameters during Stage 2. The trainable path contains 128-D WAN adapters, a 64-D action-equivalent future bottleneck, 16-D base and arm latents, and base-velocity conditioning. Adapted action-only VLA systems and runnable WAM systems provide broader context.

\paragraph{Metrics.}
Video quality is measured by frame MSE, PSNR, global SSIM~\cite{wang2004ssim}, and LPIPS~\cite{zhang2018lpips}. Action prediction is evaluated by MSE, MAE, and the mean Euclidean error of each normalized 14-D action vector. These measures jointly cover pixel fidelity, structural agreement, perceptual similarity, and control error over the complete arm--gripper--base interface used by the deployed system.

\subsection{Main Results and Reference Comparisons}

Table~\ref{tab:main_fastwam} summarizes the main FastWAM replay comparison.

\begin{center}
\centering
\refstepcounter{table}
\label{tab:main_fastwam}
{\footnotesize \textbf{TABLE~\thetable}\\
Main replay diagnostics on the 23-episode \textsc{ARMDOG} slice.\par}
\vspace{0.3em}
\scriptsize
\setlength{\tabcolsep}{3.2pt}
\begin{tabular}{@{}lrrr@{}}
\toprule
\textbf{Model} & \textbf{F-MSE$\downarrow$} & \textbf{PSNR$\uparrow$} & \textbf{A-MSE$\downarrow$} \\
\midrule
Original FastWAM (40k) & 1.616e-3 & 30.468 & 1.154e-4 \\
Original FastWAM (50k) & 1.032e-3 & 31.441 & 6.87e-5 \\
Original FastWAM (80k) & 2.136e-3 & 28.544 & 4.77e-4 \\
DECOWAM(50k) & \best{8.77e-4} & \best{31.663} & \best{5.38e-5} \\
\bottomrule
\end{tabular}
\end{center}

DECOWAM reduced frame MSE by 15.03\% and action MSE by 21.71\% relative to its 50k FastWAM initialization (Table~\ref{tab:main_fastwam}). PSNR increased by 0.222\,dB, providing a consistent pixel-space result. The 50k checkpoint is also the strongest FastWAM checkpoint in this sweep, so Stage~2 improves both prediction branches from the best observed Stage-1 starting point. The following ablations identify how the structured paths contribute to this gain.

\paragraph{Internal module ablation.}

Table~\ref{tab:internal_ablation} reports a separately trained ablation suite under the same 16-batch protocol. All variants share the Stage-1 initialization and residual adapters. Two variants remove the action-equivalent future bottleneck or ego-motion token. The adapter-only control removes all three structured paths while retaining the residual adaptation mechanism. Because this suite was trained separately, its full-model row is the appropriate within-suite reference.

\begin{center}
\centering
\refstepcounter{table}
\label{tab:internal_ablation}
{\footnotesize \textbf{TABLE~\thetable}\\
Internal ablation of the frozen decoupled FastWAM modules on the same replay protocol.\par}
\vspace{0.3em}
\scriptsize
\setlength{\tabcolsep}{1.4pt}
\begin{tabular}{@{}p{0.28\columnwidth}rrrrr@{}}
\toprule
\textbf{Variant} & \textbf{F-MSE$\downarrow$} & \textbf{PSNR$\uparrow$} & \textbf{SSIM$\uparrow$} & \textbf{A-MSE$\downarrow$} & \textbf{A-MAE$\downarrow$} \\
\midrule
Full frozen decoupled & \best{9.35e-4} & \best{31.378} & \best{9.9241e-1} & \best{8.09e-5} & \best{4.310e-3} \\
w/o quotient & 1.02e-3 & 31.228 & 9.9170e-1 & 8.40e-5 & 4.316e-3 \\
w/o base-vel. & 1.01e-3 & 31.221 & 9.9178e-1 & 8.80e-5 & 4.342e-3 \\
w/o decoupled & 9.80e-4 & 31.105 & 9.9138e-1 & 9.60e-5 & 5.135e-3 \\
\bottomrule
\end{tabular}
\end{center}

Every removal degraded all reported metrics. Relative to the adapter-only control, the full model reduced frame MSE by 4.6\%, action MSE by 15.7\%, and action MAE by 16.1\%. Removing either the future bottleneck or base-velocity conditioning worsened both output branches, demonstrating that the future-equivalent representation and explicit ego-motion signal improve the shared video--action model beyond residual adaptation alone.

\paragraph{Comparison with VLA and WAM references.}

The action-only VLA comparison isolates action quality, while the RGB column makes the additional predictive capability of the WAM interface explicit.

\begin{center}
\centering
\refstepcounter{table}
\label{tab:vla}
{\footnotesize \textbf{TABLE~\thetable}\\
Action replay comparison with VLA references on the same 23-episode \textsc{ARMDOG} slice.\par}
\vspace{0.3em}
\scriptsize
\setlength{\tabcolsep}{1.4pt}
\begin{tabular}{@{}p{0.25\columnwidth}p{0.10\columnwidth}p{0.13\columnwidth}rrr@{}}
\toprule
\textbf{Model} & \textbf{Family} & \textbf{RGB} & \textbf{A-MSE$\downarrow$} & \textbf{A-MAE$\downarrow$} & \textbf{A-L2$\downarrow$} \\
\midrule
$\pi_{0.5}$ & VLA & no & 1.79e-4 & 3.60e-3 & 2.83e-2 \\
X-VLA & VLA & no & \best{2.11e-5} & \best{1.82e-3} & \best{1.00e-2} \\
GR00T & VLA & no & 4.18e-3 & 2.51e-2 & 1.83e-1 \\
DECOWAM & WAM & yes, 8f & 5.38e-5 & 4.01e-3 & 2.24e-2 \\
\bottomrule
\end{tabular}
\end{center}

Table~\ref{tab:vla} shows the strength of a dedicated action policy: X-VLA
attains the lowest error on all three metrics. DECOWAM is nevertheless second
in A-MSE and A-L2, improving them by 69.9\% and 21.1\% over $\pi_{0.5}$; its
A-MAE is within 11.1\% of $\pi_{0.5}$ and 84.0\% lower than GR00T. Thus,
although it is not the top action-only model, DECOWAM remains in the strong VLA
range. Moreover, it simultaneously predicts a 48-step whole-body action and
eight $384\times320$ future RGB frames, whereas the VLA references stop at the
action chunk. This rollout explicitly forecasts object motion, contact geometry,
and base-induced viewpoint change, adding future awareness while retaining
competitive Action performance.

We next compare the joint interface with runnable video--action references.

\begin{center}
\centering
\refstepcounter{table}
\label{tab:wam_refs}
{\footnotesize \textbf{TABLE~\thetable}\\
World--action model references on the same \textsc{ARMDOG} data snapshot.\par}
\vspace{0.3em}
\scriptsize
\setlength{\tabcolsep}{1.2pt}
\resizebox{\columnwidth}{!}{%
\begin{tabular}{@{}lccrrrrrrr@{}}
\toprule
\textbf{Model} & \textbf{Train.} & \textbf{RGB} & \textbf{F-MSE$\downarrow$} & \textbf{PSNR$\uparrow$} & \textbf{SSIM$\uparrow$} & \textbf{LPIPS$\downarrow$} & \textbf{A-MSE$\downarrow$} & \textbf{A-MAE$\downarrow$} & \textbf{A-L2$\downarrow$} \\
\midrule
DECOWAM & 25.95M & 8f 384p & \best{8.77e-4} & \best{31.66} & \best{9.93e-1} & \best{2.95e-2} & \best{5.38e-5} & \best{4.01e-3} & \best{2.24e-2} \\
FastWAM & 6020.75M & 8f 384p & 1.03e-3 & 31.44 & 9.92e-1 & 3.03e-2 & 6.87e-5 & 4.08e-3 & 2.35e-2 \\
Motus & 5894.81M & 8f 384p & 5.19e-3 & 23.41 & 9.57e-1 & 1.01e-1 & 5.05e-4 & 9.97e-3 & 5.99e-2 \\
Cosmos 2.5 & -- & 8f 384p & 4.28e-2 & 14.38 & 6.63e-1 & 2.71e-1 & -- & -- & -- \\
X-WAM & 5037.75M & 4f 160p & 2.62e-3 & 26.57 & 9.81e-1 & 3.50e-2 & 6.31e-4 & 1.11e-2 & 2.48e-2 \\
UVA & 261.62M & 4f 128p & 1.79e-2 & 19.32 & 8.40e-1 & 2.00e-1 & 9.76e-4 & 1.31e-2 & 9.81e-2 \\
\bottomrule
\end{tabular}
}
\end{center}

Among WAM references, DECOWAM ranks first on every reported video and action
metric in Table~\ref{tab:wam_refs}. Under the matched eight-frame
$384{\times}320$ interface, it lowers FastWAM frame/action MSE by 15.03\%/21.71\%.
Relative to Motus, frame/action MSE falls by 83.1\%/89.3\%; relative to the
shorter, lower-resolution X-WAM output, the reductions are 66.6\%/91.5\%.
DECOWAM therefore adds future prediction over action-only VLAs and, unlike the
other WAMs, leads both output branches. Figure~\ref{fig:wam_reference_rollouts}
visually supports this result: DECOWAM preserves workspace geometry and object
layout while competing rollouts accumulate blur or viewpoint drift.

\begin{table*}[!t]
\centering
\refstepcounter{table}
\label{tab:real_robot}
{\footnotesize \textbf{TABLE~\thetable(a)}\\
Real-robot task progress and completion efficiency over 79 trials per method. Stage completion rates are cumulative over all attempts; best values are bold.\par}
\vspace{0.3em}
\scriptsize
\setlength{\tabcolsep}{3.2pt}
\begin{tabular*}{\textwidth}{@{\extracolsep{\fill}}lrrrrrr@{}}
\toprule
\textbf{Model} &
\shortstack{\textbf{Mean Completion}\\\textbf{Time (s) $\downarrow$}} &
\shortstack{\textbf{Cumulative Approach}\\\textbf{Completion Rate (\%) $\uparrow$}} &
\shortstack{\textbf{Cumulative Grasp}\\\textbf{Completion Rate (\%) $\uparrow$}} &
\shortstack{\textbf{Cumulative Transport}\\\textbf{Completion Rate (\%) $\uparrow$}} &
\shortstack{\textbf{Cumulative Placement}\\\textbf{Completion Rate (\%) $\uparrow$}} &
\shortstack{\textbf{Task Success}\\\textbf{Rate (\%) $\uparrow$}} \\
\midrule
GR00T & 57 & 73.4 & 13.9 & 11.4 & 8.9 & 8.9 \\
$\pi_{0.5}$ & 50 & 89.9 & 62.0 & 53.2 & 49.4 & 49.4 \\
FastWAM & 65 & 91.1 & 63.3 & 59.5 & 57.0 & 57.0 \\
X-WAM & 82 & 77.2 & 26.6 & 19.0 & 15.2 & 15.2 \\
DECOWAM & \best{49} & \best{92.4} & \best{69.6} & \best{67.1} & \best{58.2} & \best{58.2} \\
\bottomrule
\end{tabular*}
\end{table*}

\begin{center}
\centering
\refstepcounter{figure}
\label{fig:wam_reference_rollouts}
\includegraphics[width=0.98\columnwidth]{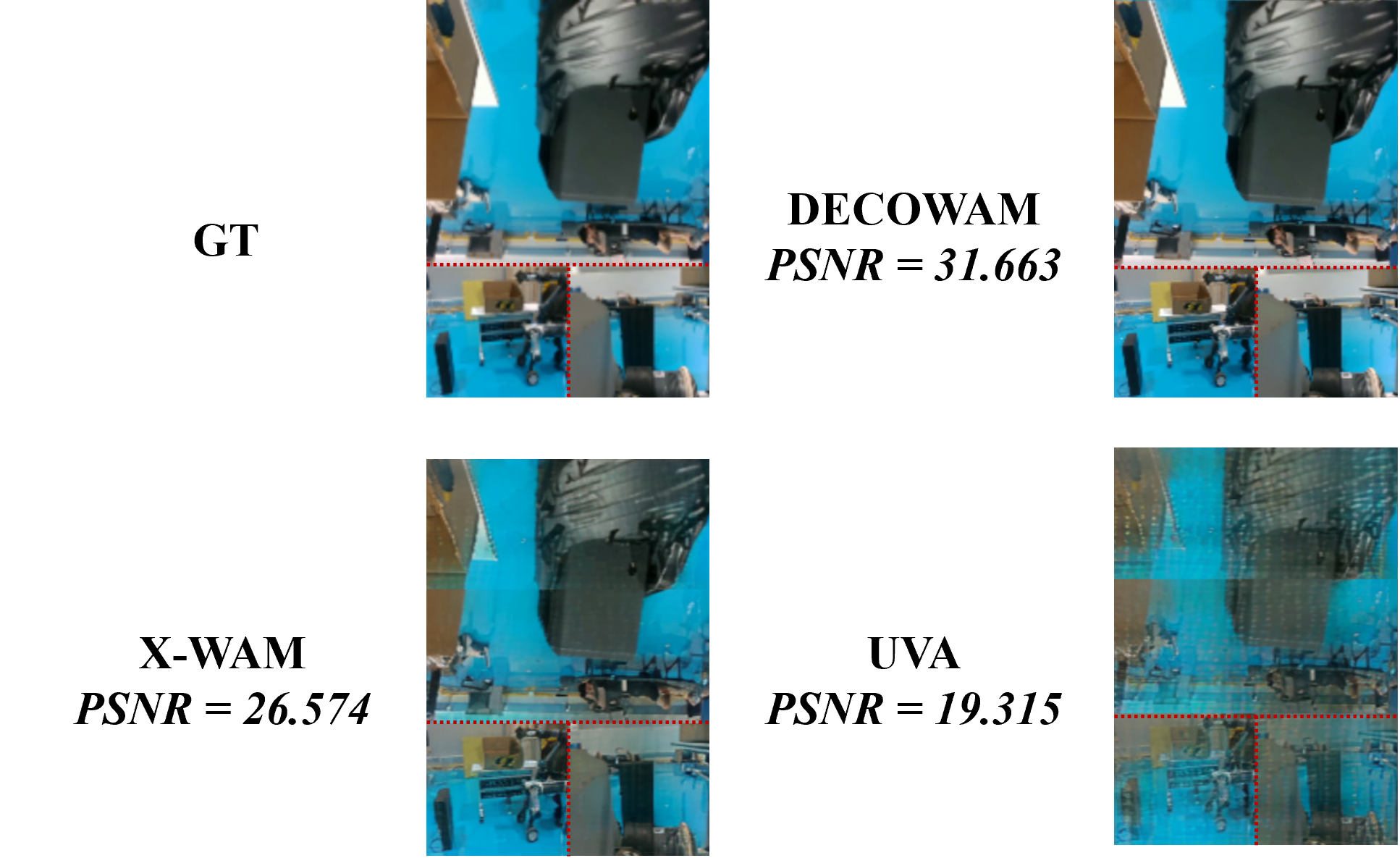}
{\footnotesize \textbf{Fig.~\thefigure.}
Qualitative comparison with WAM references on one fixed \textsc{ARMDOG} replay sample. The panels show the ground-truth future RGB montage and predictions produced through each model's native interface. Red dashed lines mark camera-view boundaries. In this example, DECOWAM better preserves workspace geometry and object layout, while several references accumulate blur or viewpoint drift.\par}
\end{center}

\subsection{Real-Robot Deployment Experiments}
\label{sec:real_robot}

Closed-loop evaluation uses the physical quadruped--arm platform with common observations, language inputs, low-level control, and safety constraints. Each method is tested in 79 trials. Table~\ref{tab:real_robot}(a) reports completion time and cumulative progress through approach, grasp, transport, placement, and task completion. Every stage uses all attempts as its denominator, revealing where failures accumulate.

Table~\ref{tab:real_robot}(b) evaluates base docking, whole-body coordination, base-displacement robustness, and autonomous recovery. Percentages are computed from counts out of 79 and rounded to one decimal place. Together, the two parts measure both end-to-end task progress and the whole-body capabilities required to sustain that progress on hardware.

DECOWAM completes 46 tasks (58.2\%) with a mean completion time of 49~s. It
records the highest approach and transport rates among the methods in
Table~\ref{tab:real_robot}(a), and carries 96.4\% of successful grasps into
transport. The table therefore confirms that the future-video branch sustains
strong closed-loop Action performance throughout the task. Among WAMs, DECOWAM
is 16~s faster than FastWAM and 33~s faster than X-WAM, while achieving 58.2\%
success versus 57.0\% and 15.2\%. Its stage profile highlights mobile-base
approach and post-grasp transport as particular strengths of decoupled
whole-body modeling.

\begin{center}
\centering
{\footnotesize \textbf{TABLE~\thetable(b)}\\
Real-robot whole-body robustness over 79 trials per method. All entries are success rates (\%); best values are bold.\par}
\vspace{0.3em}
\scriptsize
\setlength{\tabcolsep}{2.0pt}
\begin{tabular}{@{}p{0.34\columnwidth}rrrr@{}}
\toprule
\textbf{Model} & \textbf{BD-SR$\uparrow$} & \textbf{WBCM-SR$\uparrow$} & \textbf{BDP-SR$\uparrow$} & \textbf{AR-SR$\uparrow$} \\
\midrule
GR00T & 70.9 & 16.5 & 1.3 & 11.4 \\
$\pi_{0.5}$ & \best{87.3} & 36.7 & 11.4 & 25.3 \\
FastWAM & 83.5 & 34.2 & 12.7 & 27.8 \\
X-WAM & 69.6 & 27.8 & 5.1 & 12.7 \\
DECOWAM & \best{87.3} & \best{44.3} & \best{30.4} & \best{32.9} \\
\bottomrule
\end{tabular}
\end{center}

In Table~\ref{tab:real_robot}(b), BD-SR, WBCM-SR, BDP-SR, and AR-SR denote docking, coordination, displacement robustness, and recovery. DECOWAM ties the highest docking rate and leads coordination, displacement robustness, and recovery. Against FastWAM, WBCM-SR/BDP-SR rise by 10.1/17.7 points; against X-WAM, by 16.5/25.3 points. Its clearest hardware advantage is therefore sustained base--arm coordination as viewpoint and contact geometry evolve.

Relative to FastWAM, DECOWAM improves
approach, grasp, transport, and placement by 1.3, 6.3, 7.6, and 1.2 percentage
points, respectively, while reducing mean completion time by 16~s. Relative to
X-WAM, the corresponding gains are 15.2, 43.0, 48.1, and 43.0 points, together
with a 33~s reduction in completion time. The largest margins appear in
transport, whole-body coordination, and base-displacement robustness---precisely
where locomotion changes the camera viewpoint while the arm preserves a
manipulation constraint. DECOWAM therefore improves both final task completion
and the continuity of the intermediate behavior that enables it.

These real-robot results isolate the contribution of world--action modeling.
DECOWAM produces an explicit eight-frame prediction alongside its whole-body
action chunk. This performance follows from three innovations: the future
bottleneck transfers scene evolution to control, dual latents separate
navigation from manipulation, and base-velocity conditioning exposes camera
ego-motion to the video expert. Together they keep future-aware commands aligned
with the changing physical scene, extending coupled WAMs with more coherent
whole-body behavior.

\begin{figure}[t]
  \centering
  \includegraphics[width=\columnwidth]{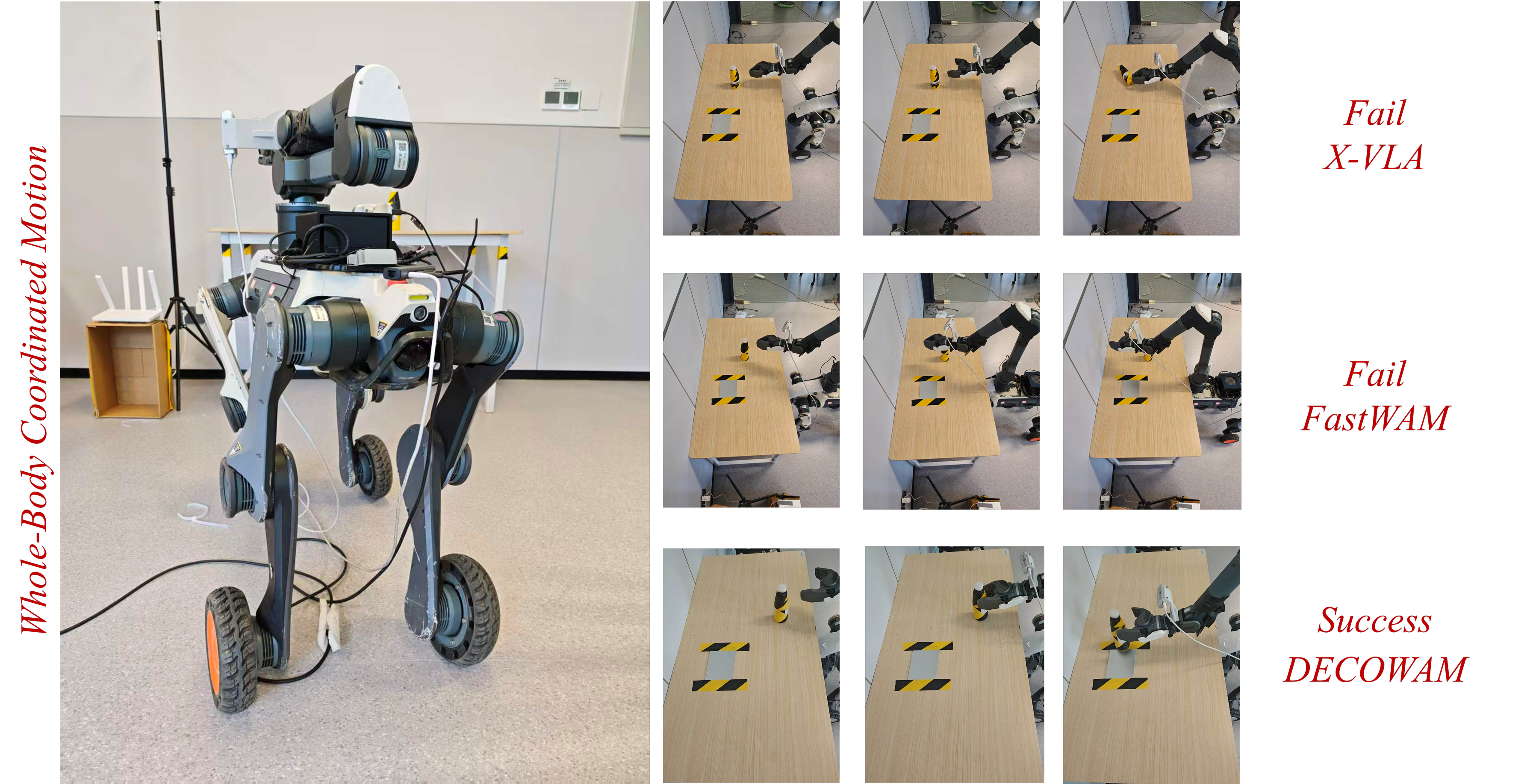}
  \caption{Representative whole-body coordinated-motion trials on the quadruped--arm platform. The three frames in each row progress from left to right. X-VLA (top) and the original FastWAM (middle) exhibit unintended base motion when a static pose is required and fail to finish the interaction, whereas DECOWAM (bottom) succeeds. The example qualitatively complements the aggregate WBCM-SR results in Table~\ref{tab:real_robot}(b).}
  \label{fig:real_robot_coordination}
\end{figure}

Figure~\ref{fig:real_robot_coordination} visualizes this advantage. X-VLA and
FastWAM fail to keep the base stationary during the tabletop reach. We attribute
this behavior to extensive training on motion-heavy data, which biases the
baseline policies toward continued movement and degrades their static-hold
capability. DECOWAM instead preserves base pose, arm reach, and end-effector
alignment to complete the interaction. Together with Table~\ref{tab:real_robot}(a),
the example shows that future-aware world--action modeling complements raw Action
accuracy by keeping commands coherent with scene evolution under whole-body motion.

\subsection{Parameter-Efficient Adaptation and Deployment}

Table~\ref{tab:runtime} separates total model size, trainable adaptation parameters, and evaluator latency. This distinction is necessary because freezing parameters reduces optimization cost without shrinking the deployed network.

\begin{center}
\begin{minipage}{\columnwidth}
\centering
\refstepcounter{table}
\label{tab:runtime}
{\footnotesize \textbf{TABLE~\thetable}\\
Deployment diagnostics on the DECOWAM checkpoint.\par}
\vspace{0.3em}
\scriptsize
\setlength{\tabcolsep}{0.8pt}
\begin{tabular}{@{}p{0.26\columnwidth}rrrrr@{}}
\toprule
\textbf{Model} & \textbf{Total M} & \textbf{Train. M} & \textbf{ms$\downarrow$} & \textbf{F-MSE$\downarrow$} & \textbf{A-MSE$\downarrow$} \\
\midrule
FastWAM & 6725.44 & 6020.75 & \best{1196.6} & 1.032e-3 & 6.87e-5 \\
DECOWAM & 6751.38 & \best{25.95} & 1333.2 & \best{8.77e-4} & \best{5.38e-5} \\
\bottomrule
\end{tabular}
\end{minipage}
\end{center}

DECOWAM reduces the number of parameters updated during Stage~2 by approximately
232-fold, from 6020.75M to 25.95M. This concentrated adaptation path improves
both frame and action MSE while adding only 11.4\% evaluator latency. Combined
with the replay and robot results, the deployment profile defines a distinct
operating point: a causal video--action interface with explicit embodiment
factors, future-frame prediction, and parameter-efficient robot specialization.

\section{Conclusion}

We presented DECOWAM, a whole-body world--action model that explicitly separates base motion, arm manipulation, and camera ego-motion. On synchronized \textsc{ARMDOG} data, staged frozen adaptation and the future-equivalent bottleneck improve every reported FastWAM video/action diagnostic while reducing Stage-2 trainable parameters from 6.021B to 25.95M. DECOWAM remains close to strong VLA Action performance while predicting eight future RGB frames, and leads the real-robot approach, transport, coordination, and displacement-robustness rates over 79 trials per method. The combined results establish a joint model that couples competitive control with explicit prediction of scene evolution under whole-body motion.

\clearpage
\begingroup
\raggedbottom
\renewcommand{\IEEEbibitemsep}{0pt}
\bibliographystyle{IEEEtran}

\endgroup

\end{document}